\documentclass[11pt]{article}
\usepackage{acl2015}
\usepackage{times}
\usepackage{url}
\usepackage{amssymb}
\usepackage{latexsym}
\usepackage{graphicx}
\usepackage{amsmath} 
\usepackage{appendix}
\usepackage{array} 
\usepackage{hyperref}
\usepackage[numbers]{natbib}
\usepackage{geometry}
\usepackage{longtable} 
\title{\LARGE The Vulnerability of Language Model Benchmarks: Do They Accurately Reflect True LLM Performance?}

\author{Sourav Banerjee \thanks {\hspace{1mm} To whom correspondence should be addressed: E-mail: sb@unitedwecare.com}\\
  DataLabs, United We Care \\\And
  Ayushi Agarwal \\
  DataLabs, United We Care \\\And
  Eishkaran Singh \\
  DataLabs, United We Care }
\date{}

\begin{document}
\maketitle
\begin{abstract}
The pursuit of leaderboard rankings in Large Language Models (LLMs) has created a fundamental paradox: models excel at standardized tests while failing to demonstrate genuine language understanding and adaptability. Our systematic analysis of NLP evaluation frameworks reveals pervasive vulnerabilities across the evaluation spectrum, from basic metrics to complex benchmarks  like GLUE and MMLU. These vulnerabilities manifest through benchmark exploitation, dataset contamination, and evaluation bias, creating a false perception of progress in language understanding capabilities. Through extensive review of contemporary evaluation approaches, we identify significant limitations in static benchmark designs, human evaluation protocols, and LLM-as-judge frameworks, all of which compromise the reliability of current performance assessments. 
As LLM capabilities evolve and existing benchmarks become redundant, we lay the groundwork for new evaluation methods that resist manipulation, minimize data contamination, and assess domain-specific tasks. This requires frameworks that are adapted dynamically, addressing current limitations and providing a more accurate reflection of LLM performance.

\end{abstract}
\section{Introduction}

% Language models, on a scale (LLMs) have brought about a change in artificial intelligence by enhancing the field of natural language processing (NLP) \cite{vaswani2017attention}. This progress has transformed the landscape of AI research and applications pushing the boundaries of what machines can accomplish in understanding and emulating abilities. As LLMs have advanced rapidly so too have the standards created to assess their performance. These benchmarks serve as a crucial yardstick in measuring the model's effectiveness across a range of linguistic tasks from generating text to tackling complex questions.\cite{devlin2019bert}, \cite{brown2020languagemodelsfewshotlearners}, \cite{radford2018improving}.

The race for state-of-the-art performance metrics in Large Language Models (LLMs) has exposed a critical vulnerability in how they are evaluated\cite{vaswani2017attention}. While LLMs achieve unprecedented scores on standardized tests, their susceptibility to dataset exploitation and pattern matching raises fundamental concerns about current evaluation methodologies. Rather than pursuing genuine language understanding, model developers increasingly exploit benchmark weaknesses, employing tactics like data contamination and climb leaderboards. Recent analyses reveal that LLMs can attain state-of-the-art results through such superficial optimization rather than deep linguistic comprehension \cite{devlin2019bert}, undermining the validity of existing benchmarking frameworks. This systematic failure in evaluation approaches threatens to misguide the field's progress, as models optimize for benchmarks that fail to capture genuine language understanding \cite{brown2020languagemodelsfewshotlearners, radford2018improving}.

\subsection{Evolution of Language Model Evaluations}
The evolution of NLP benchmarks began in the 1950s and 1960s with foundational metrics like Precision and Recall, which measure the relevance of retrieved information. The introduction of F1 Score 1979, provided a balanced evaluation by combining Precision and Recall, setting the stage for more specialized benchmarks \cite{dagan2006pascal}, \cite{sokolova2006beyond}.

Key metrics such as BLEU (2002) \cite{papineni2002bleu} for machine translation, which compares machine-generated text to human references using n-gram precision, and ROUGE (2004) \cite{lin2004rouge}, which focuses on recall in text summarization, were early attempts to evaluate language models. METEOR (2005) \cite{banerjee2005meteor} improved upon BLEU by considering synonyms and partial matches, while the RTE challenge (2005) expanded evaluation to include semantic relationships \cite{dagan2006pascal}.

The introduction of GLUE (2018) \cite{wang2019glue} marked a significant shift towards comprehensive evaluation, combining task-specific metrics across diverse NLP tasks. SuperGLUE (2019) \cite{wang2020superglue} built on this by adding more challenging tasks requiring deeper language understanding. MMLU (2021) \cite{hendrycks2021measuring} expanded evaluation to assess models across 57 subjects, though its static format limited its ability to fully capture reasoning abilities. HELM (2022) \cite{liang2023holistic} introduced a more ethically-oriented framework, evaluating models on a broader range of dimensions, including fairness and robustness. DynaBench (2022) \cite{kiela2021dynabench} pushed for dynamic, adversarial evaluation, where models are tested against human-generated challenging examples. 

The evolution of NLP benchmarks advanced with BIG-bench (2022) \cite{srivastava2023beyond}, evaluating language models across 200+ diverse tasks, and TruthfulQA (2022) \cite{lin2022truthfulqa}, which tests models' truthfulness by challenging common misconceptions. These benchmarks reflect the growing complexity and expectations in assessing modern NLP systems.

Three common themes have emerged in modern LLM testing. First, Benchmark Evaluations have expanded beyond task-specific metrics to comprehensive frameworks like GLUE, SuperGLUE, and MMLU, enabling standardized performance comparisons across diverse NLP tasks. Second, Human-as-Judge methodologies have gained prominence, particularly in challenges like DynaBench, where human annotators generate adversarial examples and assess model outputs for nuanced understanding and creativity. Third, LLM-as-Judge approaches have been introduced, utilizing other language models to evaluate outputs at scale.
As LLMs rapidly advance, the limitations of existing benchmarks have become increasingly apparent. Despite the evolution of evaluation methods, models are achieving near-perfect scores on many established metrics, raising questions about their discriminative power and comprehensiveness. This situation necessitates a critical examination of current evaluation practices and their potential vulnerabilities, especially given the high stakes associated with benchmark results in the AI landscape.
\begin{figure}
    \centering
    \includegraphics[width=\linewidth]{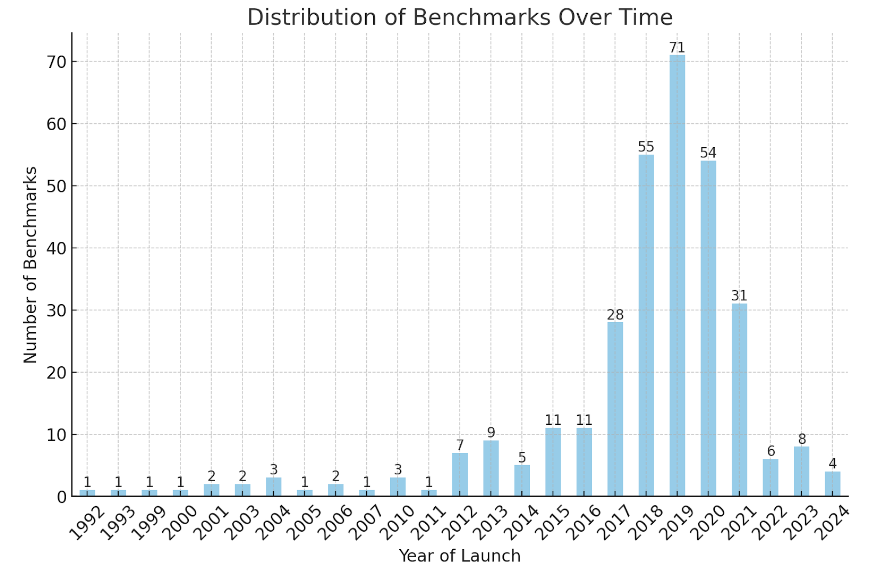}
    \caption{Benchmarks released by year till Aug 2024 \cite{openllmleaderboard}}
    \label{fig:Benchmarks released by year till Aug 2024}
\end{figure}
% \vspace{-4mm}

\section{The Vulnerability of Current Benchmarks}
The development and deployment of LLMs represent significant investments in research, computing resources, and human capital. As such, the ability to demonstrate a model's capabilities through objective measures holds considerable importance for research institutions, technology companies, and investors alike.  The introduction of new models is frequently accompanied by claims of improved performance and capabilities, and are substantiated by citing results from established benchmarks, creating a landscape where benchmark performance serves as both a measure of technological progress and a key factor in the promotion of new AI technologies \cite{chen2021evaluatinglargelanguagemodels}. $^{\ref{fig:Benchmarks released by year till Aug 2024}}$
\begin{figure}
    \centering
    \includegraphics[width=\linewidth]{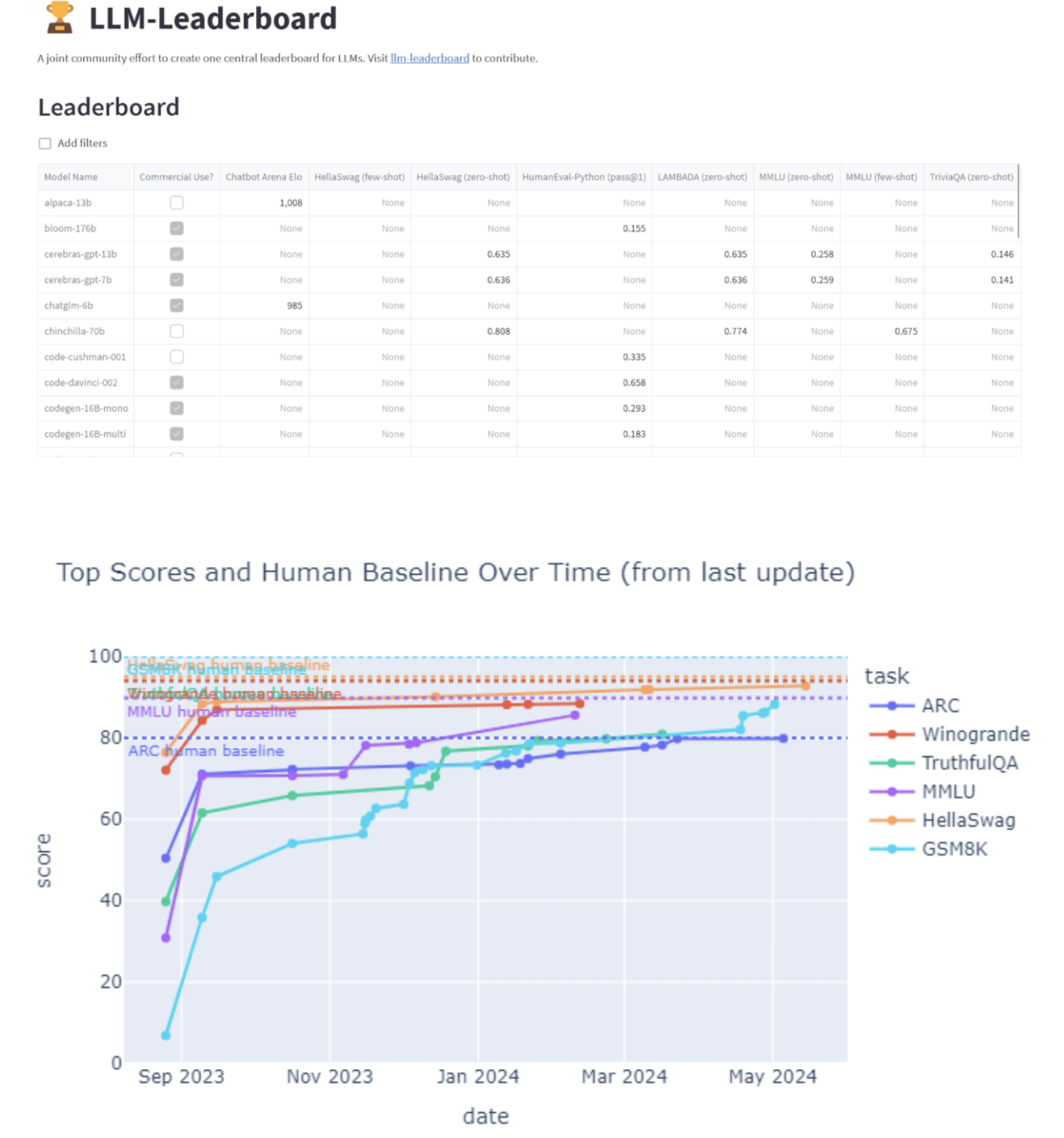}
    \caption{LLM-Leaderboard \cite{openllmleaderboard}}
    \label{fig:LLM-Leaderboard}
\end{figure}

The 'benchmark race' $^{\ref{fig:LLM-Leaderboard}}$therefore has become a hallmark of new model releases. When OpenAI introduced GPT-3, it showcased state-of-the-art results across numerous NLP tasks, including an 88.4\% score on SuperGLUE, surpassing the previous best of 84.6\% \cite{brown2020languagemodelsfewshotlearners}, \cite{wang2019glue}. Similarly, Google’s BERT set new records on eleven NLP tasks, achieving a 93.2\% F1 score on SQuAD v1.1, outperforming human performance \cite{carlini2021extracting}, \cite{wang2020superglue}. This trend has intensified with models like GPT-3.5 and GPT-4, which were highlighted for their superior performance on benchmarks, exams, and coding challenges \cite{devlin2019bert}. GPT-4, for instance, scored 86.4\% on the MMLU benchmark, well above GPT-3.5's 70.0\% \cite{hendrycks2021measuring}.

However, companies often engage in selective reporting, emphasizing their models' strengths while downplaying weaknesses . Anthropic's Claude 2, for example, focused on graduate-level exams and coding tasks \cite{radford2018improving}, while Microsoft and Nvidia’s Megatron-Turing NLG 530B highlighted its zero-shot task performance \cite{raji2019actionable}.

This scenario reflects Goodhart's Law: "When a measure becomes a target, it ceases to be a good measure" \cite{strathern1997audit}. As benchmark scores have become the primary standard for evaluating models, they have paradoxically become less reliable indicators of true capability, leading to models that excel on paper but struggle with real-world tasks requiring genuine understanding and adaptability \cite{alzahrani2024benchmarks}.

\begin{figure}
    \centering
    \includegraphics[width=\linewidth]{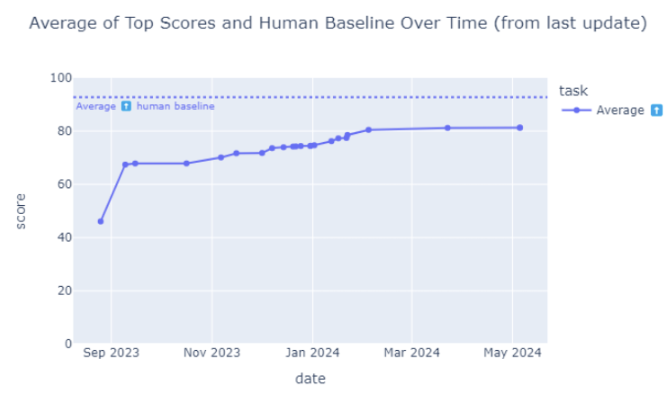}
    \caption{Open LLM Leaderboard by Hugging Face}
    \label{fig:Open LLM Leaderboard by Hugging Face}
\end{figure}

\section{How Language Model Benchmarks are “Hacked”}
The reliability of benchmarks used to evaluate Large Language Models (LLMs) is increasingly questioned due to "benchmark hacking," where \textbf{benchmarks are manipulated to exaggerate model performance.} $^{\ref{fig:LLM-Leaderboard}}$

\subsection{Benchmark Overfitting}
Benchmark overfitting, or "overtuning," occurs when models are excessively optimized for specific benchmarks without genuinely improving their general capabilities. This is akin to "p-hacking" in empirical studies, where data is manipulated to achieve misleading statistical significance \cite{brown2020languagemodelsfewshotlearners}.

To understand benchmark overfitting mathematically, let's consider a simplified model of LLM performance. Let P be the true performance of an LLM on a given task, B be the benchmark score, and $\varepsilon$ be the error term that represents the difference between the true performance and the benchmark score:
\[
B = P + \varepsilon
\]
Ideally, we want B to be an unbiased estimator of P, meaning E[B] = P, where E[] denotes expected value. However, in the case of benchmark overfitting, we often see:
\[
B>P
\]
This occurs because the model has been optimised to maximise B rather than P. We can represent this optimization process as:
\[
max(B) \neq max(P)
\]
The impact was underscored, leading to the development of more robust benchmarks like HellaSwag \cite{zellers2019hellaswag}. However, overfitting concerns persist, as seen with GPT-3's inflated performance on LAMBADA due to training data overlap and BERT's exploitation of patterns in the SQuAD dataset \cite{paperno2016lambada}, \cite{rajpurkar2018know}.

\subsection{Public Availability of Datasets}
\begin{figure}
    \centering
    \includegraphics[width=\linewidth]{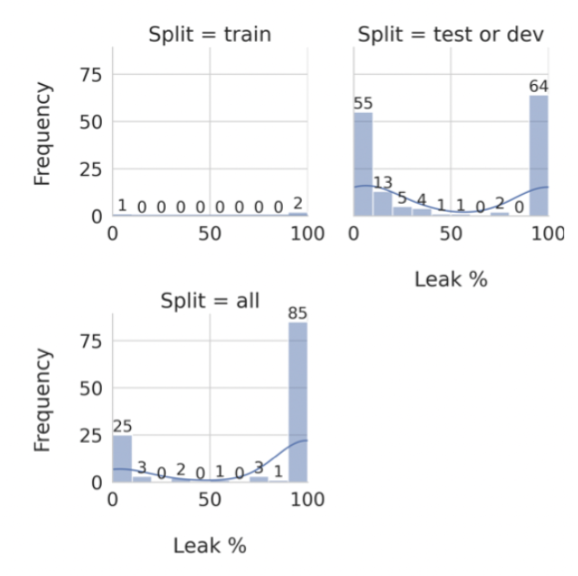}
    \caption{Data leakage distribution \cite{balloccu2024leak}}
    \label{fig:Data leakage distribution}
\end{figure}

Often referred to as "data contamination" or "test set leakage," publicly available datasets can lead to inflated performance metrics that do not accurately reflect a model's true capabilities \cite{brown2020languagemodelsfewshotlearners} $^{\ref{fig:Data leakage distribution}}$. 
Let $D_{train}$ represent the training dataset and $D_{test}$ represent the test dataset. In an ideal scenario, these sets should be disjoint:

\[ 
D_{\text{train}} \cap D_{\text{test}} = \emptyset
\]

However, due to the public nature of many benchmark datasets, we often encounter a situation where:

\[ 
D_{\text{train}} \cap D_{\text{test}} \neq \emptyset
\]

This overlap can be quantified using the contamination rate (CR):
\[
\text{CR} = \frac{|D_{\text{train}} \cap D_{\text{test}}|}{|D_{\text{test}}|}
\]

To address this issue, researchers have proposed several strategies. Dodge et al. \cite{dodge2021documenting} introduced the concept of "data contamination audits," which involve systematically checking for overlaps between training and test sets. They define a contamination score $C(m, d)$ for a model m and dataset d:
\[
C(m, d) = \sum_i s(m, x_i) \cdot |d|
\]
where $s(m, x_i)$ is a function that measures the similarity between the model's output and the ground truth for example  $x_i$. A high $C(m, d)$ suggests potential data contamination.

Another approach involves creating "hidden" test sets, kept from public release, to obtain a more accurate estimate of true model performance. However, this method faces challenges, such as ensuring the test set remains hidden and maintaining statistical validity with smaller datasets.

The implications of dataset publicity extend beyond simple memorization. LLMs trained on public datasets may develop heuristics that exploit dataset-specific patterns rather than learning generalizable language understanding. This can be conceptualised as the model learning a function g(x) that approximates the true task function $f(x)$ for
$x \in D_{test}$, but diverges for  $x \notin D_{test}$

To address these issues, the field is shifting towards more robust evaluation methods. Andrew Ilyas et al.\cite{ilyas2019adversarial}, in "Adversarial Examples Are Not Bugs, They Are Features," argue that adversarial examples exploit the very features models rely on, exposing their vulnerabilities and emphasizing the importance of testing robustness. Similarly, Robin Jia and Percy Liang's work, particularly in "Adversarial Examples for Evaluating Reading Comprehension Systems," shows that small perturbations to input data can mislead models that perform well on standard tests, highlighting the need for more rigorous evaluations. These challenge sets are designed to be more difficult and better at exposing models' weaknesses than standard tests.

\subsection{Test Set Contamination}
\begin{figure}
    \centering
    \includegraphics[width=\linewidth]{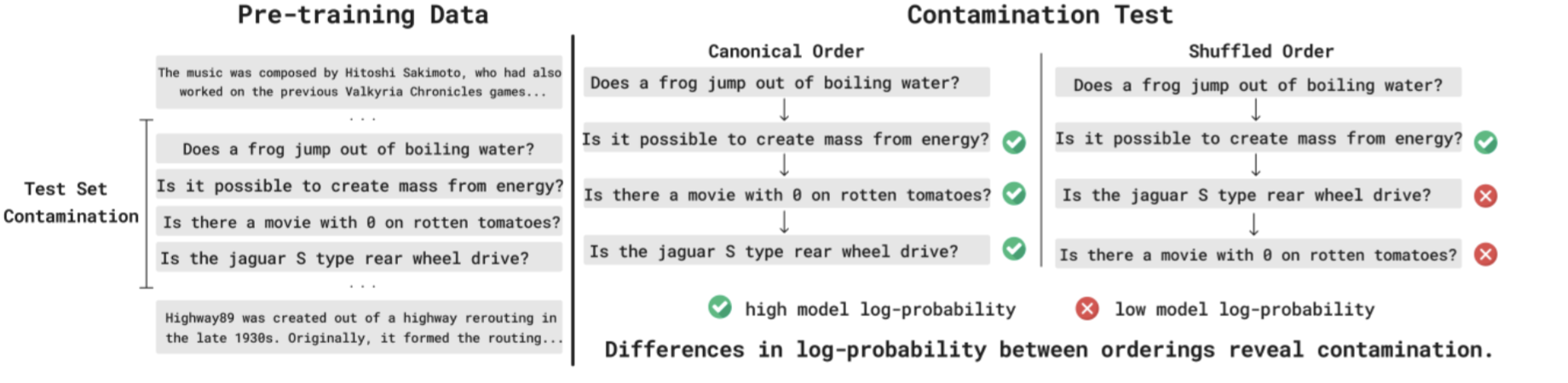}
    \caption{Detecting dataset contamination via log-probability differences in canonical vs. shuffled orders \cite{oren2023proving}}
    \label{fig:Detecting dataset contamination via log-probability}
\end{figure}

Test Set Contamination occurs when models are inadvertently exposed to test data during training, leading to skewed performance metrics and an unrealistic assessment of generalization capabilities $^{\ref{fig:Detecting dataset contamination via log-probability}}$. Let  $\theta$ represent the parameters of an LLM,  $D_{train}$ the training dataset, and $D_{test}$ the test dataset. In an ideal scenario, the model's parameters should be independent of the test set

\[
P(\theta \mid D_{\text{train}}, D_{\text{test}}) = P(\theta \mid D_{\text{train}})
\]

However, test set contamination introduces a dependency:
\[
P(\theta \mid D_{train}, D_{test}) \neq P(\theta \mid D_{train})
\]

This dependency can be quantified using mutual information 
$I(\theta; D_{test} \mid D_{train})$, which measures the reduction in uncertainty about $\theta$ given knowledge of  $D_{test}$, conditioned on $D_{train}$ \cite{carlini2021extracting}:

\[
I(\theta; D_{\text{test}} \mid D_{\text{train}}) = H(\theta \mid D_{\text{train}}) - H(\theta \mid D_{\text{train}}, D_{\text{test}})
\]

where H() denotes entropy. A non-zero value of  $I(\theta; D_{test} \mid D_{train})$ indicates potential test set contamination.
The severity of contamination can vary. In some cases, entire test examples may be present in the training data, a situation we can represent as:
\[
\exists x \in D_{\text{test}} : x \in D_{\text{train}}
\]
More subtle forms of contamination can occur when the training data contains information that is highly correlated with the test set. This can be modelled using a similarity function $sim(x, y)$ between examples:
\[
\exists x \in D_{\text{train}}, y \in D_{\text{test}} : \text{sim}(x, y) > \tau 
\]
where $\tau$ is some threshold of similarity.

This contamination can lead to inflated performance estimates, where the model's generalization function f(x)  is conflated with a contamination-influenced function g(x) :

\[
g(x) = f(x) + c(x)\]

where c(x) represents the contribution from test set contamination.

To quantify the impact of contamination, researchers have proposed various metrics. Carlini et al. \cite{carlini2019secret} introduced the "exposure" metric, which measures how much information about a specific training example is encoded in the model's parameters. They define exposure as:
\[
\text{Exposure}(s[r]) = \log_2 |R| - \log_2(\text{rank}(s[r]))
\]

where $|R|$ is the size of the randomness space from which the canaries (random sequences) are chosen, and $rank_{\theta}(s[r])$ is the rank of a specific canary sequence $s[r]$ based on the model's perplexity, where the rank represents the index of the canary in the list of all possible canaries sorted by their log-perplexity. High exposure values indicate potential memorization and contamination.

To mitigate contamination, several strategies have emerged, including dynamic test set generation, where test data is created post-training to avoid overlaps, and data provenance tracking \cite{kale2023provenance}, which involves tracing the origin and history of each data example. Provenance tracking helps assess contamination risk by comparing test and training data histories, flagging high-risk examples for removal or further investigation \cite{nie2020adversarial}.

Future work may involve combining multiple strategies, such as dynamic test set generation, rigorous provenance tracking, and advanced contamination detection techniques, to ensure that reported model performance accurately reflects true generalization capabilities.

\subsection{Gaming Through Task-Specific Optimization}
Language model developers often exploit specific benchmark structures to achieve high performance scores, a practice that may yield impressive results on certain tasks but fails to enhance overall model capabilities or generalization to diverse contexts \cite{brown2020languagemodelsfewshotlearners}. 

Let $M$ be an LLM with parameters $\theta$, and $B = {B_1, B_2, ..., B_n}$ be a set of benchmark tasks. The optimization problem for task-specific fine-tuning can be formulated as:
\[
\arg\max_{\theta} \sum_{i} w_{i} S(M_{\theta}, B_{i})
\]
where $S(M, B_{i})$ is the score of model $M$ with parameters $\theta$ on benchmark $B_{i}$, and $w_{i}$ are task-specific weights. This formulation encourages overfitting to the specific characteristics of the chosen benchmarks rather than improving general language understanding.

The discrepancy between benchmark performance and real-world capability can be quantified using a generalization gap metric:
\[
G(M) = \mathbb{E}[S(M, B_{\text{test}})] - \mathbb{E}[S(M, B_{\text{real}})]
\]

where $B_{test}$ represents benchmark test sets and $B_{real}$ represents real-world tasks. A large $G(M)$ indicates potential gaming of the benchmarks.

McCoy demonstrated this issue with BERT models fine-tuned on the MNLI dataset \cite{williams2018broadcoverage}. They showed that these models often relied on shallow heuristics specific to the dataset structure rather than developing a deeper understanding of natural language inference. For instance, models achieved high accuracy by exploiting the presence of specific words or phrase structures that were correlated with certain labels in the training data.This can be represented as a learned decision function $f(x)$ that approximates the true task function $t(x)$ for $x \in B_{test}$, but diverges for $x \notin B_{test}$:
\[
\|f(x) - t(x)\| \approx 0 \text{ for } x \in B_{\text{test}}
\]
\[
\|f(x) - t(x)\| \gg 0 \text{ for } x \notin B_{\text{test}}
\]
where $\|\cdot\|$ denotes an appropriate distance metric in the output space.

Similarly, Kaushik et al. \cite{kaushik2020learning} found that models trained on the SNLI dataset often relied on artifact features rather than genuine semantic understanding. They proposed a counterfactually-augmented data approach to mitigate this issue, effectively expanding the task distribution to reduce exploitable patterns:
\[
B'_{\text{train}} = B_{\text{train}} \cup \{ c(x, y) \mid (x, y) \in B_{\text{train}} \}
\]
where $c(x, y)$ generates a counterfactual example for input  $x$ and label $y$.
To counteract these problems, researchers have suggested creating more challenging benchmarks, like SuperGLUE \cite{wang2020superglue}, that require complex reasoning, and employing multi-task learning to avoid overfitting. Adversarial testing methods, such as the Adversarial NLI dataset, have also been explored to expose the limitations of task-specific optimization.

To mitigate these issues, the field is moving towards more holistic evaluation frameworks. For instance, Bowman and Dahl \cite{bowman2021fixing} advocate for a shift from narrowly-defined leaderboard tasks to more comprehensive evaluation suites that assess a wide range of linguistic and reasoning capabilities. Future research directions may involve the development of more robust, multi-faceted evaluation frameworks, the creation of benchmarks that are inherently resistant to gaming, and a greater emphasis on assessing model performance in diverse, real-world contexts.

\subsection{Adversarial Benchmarking}
Adversarial benchmarking has become crucial for assessing the robustness of large language models (LLMs) against subtle, malicious inputs designed to confuse or mislead them \cite{brown2020languagemodelsfewshotlearners}. Traditional benchmarks often fail to capture these vulnerabilities, leading to overly optimistic assessments of model performance \cite{xhonneux2024efficient}. Let M be an LLM with parameters $\theta$, and $x$ be an input from the distribution D. The model's output can be represented as:
\[
y=M_{\theta}(x)
\]
An adversarial example x' can be defined as:
\[
x' = x + \delta, \quad \text{such that } \|\delta\| \leq \epsilon \text{ and } M_\theta(x') \neq M_\theta(x)
\]

where $\| \cdot \|$ is some distance metric and $\epsilon$ is a small perturbation bound. The goal of adversarial benchmarking is to evaluate the model's performance on these perturbed inputs:
\[
R(M) = \mathbb{E}_x[L(M(x'), y)]
\]

where L is a loss function and y is the true label. A lower R(M) indicates higher robustness.

Jia and Liang \cite{rajpurkar2018know} demonstrated the vulnerability of reading comprehension systems to adversarial inputs by introducing the AddSent method. They showed that appending adversarially constructed sentences to paragraphs could dramatically reduce the performance of state-of-the-art models on the SQuAD dataset. This can be formalized as:
\[
x' = x \oplus s 
\]
where $\oplus$ denotes concatenation and s is an adversarially generated sentence. They found that:
\[
E[S(M, x')] \ll E[S(M, x)]
\]
where S is the model's performance score.

Similarly, Alzantot et al. \cite{alzantot2018generating} introduced a genetic algorithm-based approach to generate adversarial examples for text classification tasks. Their method iteratively applies small perturbations to inputs while maintaining semantic similarity and grammatical correctness:
\[
x_{t+1}' = \arg\max_{x'} \{ f(x') \mid \text{sim}(x', x) > \tau, \, g(x') = 1 \}
\]
where $f$ is the adversarial objective, sim is a semantic similarity function, $\tau$ is a threshold, and g is a grammaticality checker. They demonstrated that models achieving high accuracy on standard benchmarks could be easily fooled by these adversarial inputs.

The implications are significant; models that excel on standard benchmarks may fail under adversarial conditions in real-world scenarios. To quantify this, metrics like Adversarial Accuracy Drop (AAD) measure the difference in model performance between original and adversarial datasets \cite{olivier2023how}, \cite{balaji2019instance}. High AAD indicates vulnerability to adversarial attacks.

Adversarial benchmarks have revealed surprising weaknesses in models like BERT, which Wallace et al. \cite{raji2019actionable} showed to be vulnerable to universal adversarial triggers—short token sequences that consistently cause specific mispredictions. To improve robustness, techniques like FreeLB \cite{zhu2020freelb}, add adversarial perturbations during training, enhancing model generalization across tasks.

To counter these weaknesses, adversarial benchmarking initiatives have emerged. Nie et al. [4] developed the Adversarial NLI (ANLI) dataset, iteratively designed to challenge models as they improve, while Gardner et al. \cite{wallace2019universal} proposed contrast sets—minimally edited benchmark examples that expose model vulnerabilities.

The rise of adversarial benchmarks has also driven research into model interpretability and analysis. For example, Gururangan et al. \cite{gururangan2018annotation} used adversarial examples to uncover annotation artifacts in natural language inference datasets, leading to improved data collection practices. Future research may focus on dynamic adversarial benchmarks that evolve with model improvements, multi-modal adversarial inputs, and integrating adversarial robustness as a core metric in model evaluation.

\subsection{Human Bias }
\begin{figure}
    \centering
    \includegraphics[width=\linewidth]{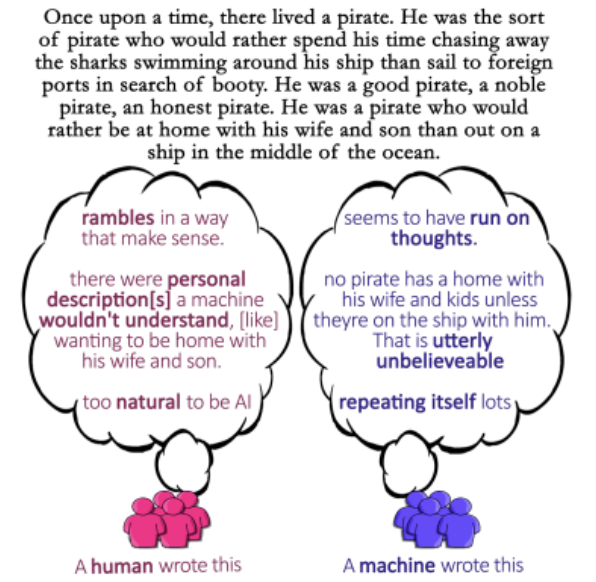}
    \caption{Excerpts from human evaluators’ explanations for why they believe a GPT3-generated story (also excerpted) was written by a human (left) or a machine (right). The evaluators point to a wide range of text attributes to make their decisions, sometimes using the same aspect of the text to come to opposite conclusions \cite{clark2021all}}
    \label{fig:Human-Bias}
\end{figure}

The increasing reliance on human judges to evaluate the performance of large language models, exemplified by approaches like DynaBench \cite{kiela2021dynabench} that emphasize adversarial human examples, has introduced new vulnerabilities in the benchmarking process. While human judgment provides valuable insights into model capabilities, it also opens the door to potential benchmark hacking. One significant limitation of human evaluations is the inherent inconsistency and bias related to factors like output formatting, tone, and formality \cite{chiang2024chatbot}. Oren et al. (2023) highlight that different human annotators may apply varying standards when assessing model outputs, resulting in noisy and unreliable scores \cite{oren2023proving}. This lack of consistency can be exploited by models that are finely tuned to the specific preferences and idiosyncrasies of individual human judges, rather than focusing on genuine improvements in language understanding. Additionally, human participants may not create a diverse range of questions, may focus on certain topics that do not thoroughly test a model's overall abilities, or may design prompts that are poorly constructed \cite{zheng2024judging}.$^{\ref{fig:Human-Bias}}$

One key limitation of human evaluations is the inherent inconsistency and bias related to factors like output formatting, tone, and formality \cite{chiang2024chatbot}. Oren et al. (2023) highlight that different human annotators may apply varying standards when assessing model outputs, resulting in noisy and unreliable scores \cite{oren2023proving}. This lack of consistency can be exploited by models that are finely tuned to the specific preferences and idiosyncrasies of individual human judges, rather than focusing on genuine improvements in language understanding. Furthermore, human participants may not create a diverse range of questions, may concentrate on certain topics that do not thoroughly test a model's overall abilities, or may design prompts that are poorly constructed \cite{zheng2024judging}.
Moreover, human evaluations are susceptible to adversarial attacks specifically designed to deceive the judges. Oren et al. (2023) demonstrate that LLMs can generate outputs crafted to appear convincing to humans, even when the content is factually incorrect or logically inconsistent \cite{oren2023proving}. By exploiting human cognitive biases and heuristics, models can manipulate benchmark scores without exhibiting true language understanding. This vulnerability underscores the need for more robust evaluation methods that can detect and mitigate adversarial gaming strategies.

The use of human judges can also create perverse incentives for models to prioritize surface-level features that humans find persuasive, such as coherence and fluency, at the expense of deeper aspects of language comprehension. Dubois et al. (2024) argue that this phenomenon, termed "evaluation hacking," leads models to optimize for human-pleasing outputs rather than developing genuine linguistic competence \cite{dubois2024lengthcontrolled}. Furthering this concern, Clark et al. (2021) illustrate the subjective nature of human evaluations in discerning between human and machine-generated text. Their study reveals evaluators' contradictory judgments, often influenced by the same superficial features, such as coherence, that models are programmed to exploit. This inconsistency highlights the unreliability of human assessments and the necessity for more objective evaluation methods that prioritize genuine linguistic competence over appealing yet shallow textual attributes \cite{clark2021all}.

To address these challenges, researchers have proposed various mitigation strategies. One approach is to employ adversarial filtering techniques to identify and exclude human-generated examples that are designed to deceive models \cite{oren2023proving}. By proactively removing adversarial inputs from the evaluation set, the impact of gaming strategies can be reduced. Additionally, the use of multiple diverse human judges, rather than relying on a single evaluator, can help to average out individual biases and provide a more balanced assessment of model performance \cite{dubois2024lengthcontrolled}. However, while these techniques can alleviate some of the issues associated with human judging, the inherent subjectivity and potential for gaming remain significant obstacles to fully reliable benchmarking.

\subsection{Model Overfitting to Peer Evaluation}
As the limitations of human judges became apparent, researchers explored the use of large language models (LLMs) themselves as evaluators for other language models. By leveraging the capabilities of LLMs to automatically assess the outputs of target models, this approach aimed to scale up evaluation and reduce the influence of human biases. However, the use of LLM-based judging introduces its own set of vulnerabilities that can enable benchmark hacking.

One major concern with LLMs as judges is the potential for model-to-model collusion. Yan (2024) demonstrates that LLMs can learn to identify and exploit regularities in the outputs of other models, leading to inflated performance estimates \cite{yan2024taskspecific}. By detecting patterns that are specific to certain model families or architectures, LLMs can assign high scores to outputs that match these patterns, regardless of their actual quality or relevance to the task at hand. Furthermore, researchers have found that LLMs often favor their own answers over those of other LLMs, and they tend to prefer more verbose outputs \cite{dubois2024length}, \cite{li2024arena}, \cite{li2023alpacaeval}, \cite{white2024livebench}.

The use of LLMs as judges can also amplify and perpetuate biases present in the training data of both the judge and target models. If an LLM judge has internalized certain biases or learned to favor specific types of outputs based on spurious correlations in its training data, it may assign disproportionately high scores to target models that exhibit similar biases \cite{yan2024taskspecific}. This can lead to a distorted view of model performance, as the benchmarks fail to accurately assess the models' ability to generate unbiased and equitable outputs. The amplification of biases through LLM-based judging can have significant implications, particularly in sensitive domains such as hate speech detection or content moderation, where the fair treatment of all users is of utmost importance.

Another risk associated with using LLMs as judges is the potential to stifle innovation and hinder the discovery of novel approaches to language modeling. If an LLM judge has been trained on a narrow range of model architectures or has developed biases towards the patterns and behaviors of models similar to itself, it may assign lower scores to outputs generated by radically different architectures, even if these new approaches represent genuine advances in the field \cite{yan2024taskspecific}. This bias towards familiar patterns can create a self-reinforcing cycle, where established model architectures are favored over innovative techniques, slowing down progress and limiting the exploration of alternative solutions.

To mitigate these challenges, researchers have proposed various strategies for ensuring fair and unbiased evaluation with LLM judges. Yan (2024) suggests the use of "judge ensembles," which aggregate scores from a diverse set of LLM architectures to reduce the impact of model-specific biases \cite{yan2024taskspecific}. By combining the assessments of multiple LLMs with different training backgrounds and architectures, the influence of individual model biases can be diluted, resulting in a more balanced evaluation. Additionally, the development of techniques for auditing LLM judges for biases and ensuring their fairness is crucial to maintain the integrity of benchmarking results.

However, despite these mitigation efforts, the fundamental similarities between LLM judges and the target models they evaluate remain a significant concern. As long as both the judge and target models are based on similar architectures and training paradigms, there is an inherent risk of shared biases and blind spots that can undermine the reliability of benchmarking results. Overcoming these challenges will require a concerted effort from the research community to develop novel evaluation approaches that are truly independent of the models being assessed and can provide a more objective measure of language understanding capabilities.

The vulnerabilities introduced by both human and LLM judges, including the sensitivity of LLM evaluation to prompting effects, serve as a stark reminder of the ongoing challenges in creating robust and reliable benchmarks for evaluating language model performance. As the field of natural language processing continues to evolve at a rapid pace, it is imperative that researchers remain vigilant to the potential for benchmark hacking and actively work to develop evaluation methodologies that are resilient to gaming and manipulation. This may involve a combination of techniques, such as adversarial filtering, bias auditing, and the use of diverse judge ensembles, as well as the exploration of entire

\subsection{Human Evaluation: A Flawed Gold Standard}
Human evaluation has long been considered the gold standard for assessing language model outputs. However, Clark et al. demonstrate in their study that human evaluations of generated text are fraught with inconsistencies and biases \cite{clark2021all}. In their analysis of GPT-3 generated stories, they found that human evaluators often relied on superficial text attributes to make their judgments, leading to contradictory assessments.

For instance, evaluators frequently cited "coherence" as a reason for both human and machine attributions. One evaluator noted, "The story flows well and is coherent, which suggests it was written by a human," while another stated, "The text is too coherent and flows too smoothly to be written by a human." This contradiction highlights the subjective nature of human evaluation and its vulnerability to individual biases and expectations.

Furthermore, the study revealed that evaluators were often incorrect in their attributions, with accuracy rates barely above chance (51.9\% to 72.3\%) across various experimental conditions \cite{clark2021all}. These findings have significant implications for benchmark evaluation. The inconsistency in human judgments calls into question the reliability of human-evaluated benchmarks. 

Additionally, LLMs trained on human-evaluated datasets may learn to exploit these superficial attributes, potentially leading to models that appear more "human-like" without genuine improvements in language understanding. There's also a risk that models could be fine-tuned to produce outputs that cater to known human biases, artificially inflating their scores on human-evaluated benchmarks.

\subsection{LLMs as Judges: A Double-Edged Sword}

As an alternative to human evaluation, recent research has explored the use of LLMs themselves as judges for evaluating other language models. Yin et al. conducted a comprehensive study on this approach, revealing both promising aspects and significant limitations \cite{chiang2023can}. Their findings provide a nuanced view of the potential and pitfalls of using LLMs as judges.

The study found several advantages to the LLM-as-judge approach. LLMs demonstrated higher inter-annotator agreement compared to human evaluators, with GPT-4 achieving a Fleiss' kappa of 0.686 versus 0.270 for humans on the WMT dataset. LLM evaluations also showed strong correlations with aggregated human scores, with GPT-4 achieving Pearson correlations of 0.81 and 0.74 on the WMT and GEM datasets, respectively. Furthermore, LLM-based evaluation can be conducted more rapidly and at a larger scale compared to human evaluation, potentially allowing for more comprehensive benchmarking \cite{chiang2023can}.

However, the study also uncovered several critical limitations. LLM judgments were highly sensitive to prompt formulation, with even minor changes leading to significant shifts in evaluation outcomes. While LLMs could provide explanations for their judgments, these explanations were often inconsistent with the actual scores given, raising questions about the interpretability and reliability of their decision-making process. The researchers also found that LLMs consistently favored longer, more detailed responses, potentially leading to inflated scores for prolific but not necessarily higher-quality outputs. Perhaps most concerningly, the study demonstrated that LLM judges could be manipulated by carefully crafted inputs designed to exploit their biases or limitations \cite{chiang2023can}.

Further developments, such as the Agent-as-a-Judge framework introduced by Zhuge et al\cite{zhuge2024agentasajudgeevaluateagentsagents}, aim to address these limitations by incorporating iterative feedback loops, where agents assess one another throughout multi-step tasks. This agent-based approach, validated on benchmarks like DevAI, enhances evaluation precision and adaptability by integrating real-time intermediate assessments, suggesting a pathway toward mitigating biases inherent in traditional LLM judgment structures.

\subsection{Systematic Vulnerabilities in LLM-as-Judge Evaluations}
Building on these findings, Zheng et al. conducted a more focused study on the vulnerabilities of LLM-as-judge evaluations \cite{zheng2024judging}. Their research uncovered several systemic issues that could lead to misleading benchmark results.

The study found a significant self-preference bias, where LLMs consistently rated their own outputs higher than those of other models, including human-written responses. This bias was observed across multiple LLMs and tasks, with an average self-preference of 55.80\% for instruction-following tasks and 60.40\% for open-ended tasks. The researchers also noted instability across LLMs, with different LLM judges producing inconsistent rankings of the same set of models. The average Kendall's Tau correlation between judge pairs was only 0.582, indicating a concerning lack of consensus \cite{zheng2024judging}.

Furthermore, the study demonstrated that models could be fine-tuned to exploit known biases of LLM judges, artificially inflating their benchmark scores without genuine improvements in capability. LLM judges also showed high sensitivity to superficial changes in input formatting and phrasing, leading to significant fluctuations in scores for essentially identical content \cite{zheng2024judging}.
These vulnerabilities raise serious concerns about the validity of LLM-as-judge evaluations and their potential to produce misleading benchmark results. Models could be optimized to perform well under specific LLM judges without necessarily improving their overall language understanding or task performance.

\subsection{Implications for Benchmark Interpretation}
The findings from these studies have significant implications for how we interpret and use LLM benchmark results. As models approach or achieve perfect scores on established benchmarks, these results should be viewed with increased skepticism, considering the potential for benchmark hacking or exploitation of evaluation biases.

No single evaluation method—whether human, LLM-as-judge, or static metrics—can provide a comprehensive assessment of LLM performance. A combination of approaches, each with clearly stated limitations, is necessary for a more robust evaluation. As models become more adept at exploiting current evaluation methods, there is a pressing need for benchmarks to evolve continuously, introducing novel tasks and evaluation criteria that challenge models in unexpected ways.

Researchers and organizations should provide detailed information about their evaluation methodologies, including potential biases and limitations, to allow for more informed interpretation of benchmark results. While benchmarks provide valuable comparative data, increased emphasis should be placed on evaluating LLMs in diverse, real-world applications where the complexity and unpredictability of tasks may reveal capabilities or limitations not captured by current benchmarks.

While LLM benchmarks remain an important tool for measuring progress in natural language processing, their vulnerabilities to gaming and bias necessitate a more nuanced and critical approach to result interpretation. As the field continues to advance, developing more robust, dynamic, and comprehensive evaluation frameworks will be crucial to ensure that benchmark results accurately reflect genuine improvements in language model capabilities.

\section{Analysis of Systematic Benchmark Manipulation in Language Models}
The evaluation of Large Language Models (LLMs) through benchmarking presents significant methodological challenges that warrant critical examination. Recent investigations reveal systematic patterns in benchmark optimization that may compromise the validity of reported performance metrics. Contemporary benchmark methodologies often fail to capture the full spectrum of model performance, instead providing potentially misleading indicators of capability. This phenomenon manifests through multiple vectors: structural biases in benchmark design, statistical validity concerns, resource distribution inequities, standardization limitations, and selective reporting practices. Each of these factors contributes to a complex ecosystem where benchmark results may not accurately reflect true model capabilities or generalization potential.
\subsection{Structural Biases in Benchmark Design}
Current benchmark methodologies exhibit inherent structural biases that potentially distort performance metrics. For instance, the GPT-3 evaluation protocol demonstrates performance optimization through selective task curation, particularly in few-shot learning scenarios \cite{brown2020languagemodelsfewshotlearners}. While such evaluations yield quantifiable results, they may not adequately reflect the model's generalized performance across diverse application contexts.
\subsection{Statistical Validity and Sample Size Considerations}
The evaluation of LLM performance metrics necessitates careful consideration of statistical principles, particularly the Law of Large Numbers. Statistical inference based on limited samples may yield unreliable conclusions regarding model capabilities. This phenomenon is exemplified in the evaluation of the LLaMA model \cite{touvron2023llama}, which reports competitive performance metrics despite utilizing a significantly reduced training dataset compared to contemporary models. The statistical reliability of such results warrants scrutiny, as performance metrics derived from limited datasets may not accurately represent the model's general capabilities. This statistical uncertainty parallels fundamental probability theory, where limited sampling can produce deceptive apparent probabilities that fail to reflect true underlying distributions.
\subsection{Resource Distribution and Historical Advantages in Model Development}
The development and evaluation of LLMs exhibits significant resource-based stratification within the field. Established organizations with extensive computational infrastructure and historical data accumulation, such as those developing GPT-4 \cite{openai2024gpt4}, possess substantial advantages in model optimization and performance refinement. This resource asymmetry manifests in multiple dimensions: access to extensive historical datasets, accumulated technical expertise, and refined fine-tuning methodologies developed over extended periods. These advantages create systemic barriers to equitable model comparison, particularly for emerging research entities with limited access to comparable resources.

The impact of this resource disparity extends beyond mere computational capability. Organizations with extensive development histories benefit from accumulated technical knowledge, refined training methodologies, and extensive proprietary datasets. This historical advantage creates inherent biases in benchmark performance metrics, potentially obscuring the actual algorithmic or architectural innovations of newer models. The phenomenon necessitates careful consideration in benchmark design to ensure fair evaluation of fundamental model capabilities independent of resource advantages.

\subsection{Limitations of Standardized Evaluation Protocols}
Current standardized evaluation frameworks, such as MMLU \cite{hendrycks2021measuring}, which assesses models across 57 domains, present methodological constraints in capturing the full spectrum of language understanding and generation capabilities. These protocols may inadequately measure adaptive reasoning and contextual understanding, particularly in dynamic, real-world applications.
\subsection{Selective Reporting in Performance Analysis}
The documentation and reporting of benchmark results frequently exhibit systematic biases in data presentation. For instance, the Qwen model documentation \cite{bai2023qwen} emphasizes performance metrics in specific linguistic domains while potentially understating performance variations in others. This selective reporting methodology necessitates more comprehensive and standardized documentation protocols.

This analysis reveals systematic limitations in current benchmark methodologies, suggesting the need for more rigorous evaluation frameworks that address these identified constraints. The development of a zero-day, zero-shot evaluation framework represents a methodological response to these challenges, introducing protocols designed to address these identified limitations while upholding evaluation rigor. 

\section{Future Work}
Future research proposes substantive methodological improvements to address the identified limitations in current LLM evaluation frameworks. We propose, a zero-day, zero-shot evaluation protocols initially centered on specific domain applications. We aim to introduce systematic assessment mechanisms that subject language models to novel, domain-specific scenarios, thereby quantifying their contextual understanding and adaptive reasoning capabilities.

We will explore dynamic framework for evaluating large language models (LLMs), underpinned by consortium-based governance structure comprising representatives from academia, industry, and research organizations. The proposed governance mechanism will ensure systematic quarterly iterations to mitigate benchmark decay and prevent overfitting, fostering continuous improvement. We also propose parameter-normalized peer review systems, which use standardized protocols to evaluate responses anonymously, ensures fairness and consistency across computational architectures.

As a proposed approach for future research, we aim to incorporate domain-specific adaptations to ensure that tasks in legal, financial, and educational sectors are thoroughly tested. We plan to establish domain-specific evaluation protocols while maintaining methodological rigor, integrating zero-day assessment protocols, structured governance, and dynamic iteration processes. This strategy marks a shift from static benchmarks to adaptive evaluation mechanisms. By emphasizing domain expertise and consistent methodological evolution, we intend to provide a robust foundation for advancing LLM capabilities and ensuring their reliability in high-stakes, real-world contexts.

\bibliographystyle{acl}

\begin{thebibliography}{}

\bibitem[1]{brown2020languagemodelsfewshotlearners}
Tom B. Brown, Benjamin Mann, Nick Ryder, Melanie Subbiah, Jared Kaplan, Prafulla Dhariwal, Arvind Neelakantan, Pranav Shyam, Girish Sastry, Amanda Askell, Sandhini Agarwal, Ariel Herbert-Voss, Gretchen Krueger, Tom Henighan, Rewon Child, Aditya Ramesh, Daniel M. Ziegler, Jeffrey Wu, Clemens Winter, Christopher Hesse, Mark Chen, Eric Sigler, Mateusz Litwin, Scott Gray, Benjamin Chess, Jack Clark, Christopher Berner, Sam McCandlish, Alec Radford, Ilya Sutskever, and Dario Amodei.
2020.
Language Models are Few-Shot Learners.

\bibitem[2]{carlini2021extracting}
Nicholas Carlini, Florian Tram{\`e}r, Eric Wallace, Matthew Jagielski, Ariel Herbert-Voss, Katherine Lee, Adam Roberts, Tom Brown, Dawn Song, {\'U}lfar Erlingsson, Alina Oprea, and Colin Raffel.
2021.
Extracting Training Data from Large Language Models.
In \textit{30th USENIX Security Symposium (USENIX Security 21)}, pages 2633--2650. USENIX Association.

\bibitem[3]{devlin2019bert}
Jacob Devlin, Ming-Wei Chang, Kenton Lee, and Kristina Toutanova.
2019.
BERT: Pre-training of Deep Bidirectional Transformers for Language Understanding.
In \textit{Proceedings of the 2019 Conference of the North American Chapter of the Association for Computational Linguistics: Human Language Technologies, Volume 1 (Long and Short Papers)}, pages 4171--4186. Association for Computational Linguistics.

\bibitem[4]{chen2021evaluatinglargelanguagemodels}
Mark Chen, Jerry Tworek, Heewoo Jun, Qiming Yuan, Henrique Ponde de Oliveira Pinto, Jared Kaplan, Harri Edwards, Yuri Burda, Nicholas Joseph, Greg Brockman, Alex Ray, Raul Puri, Gretchen Krueger, Michael Petrov, Heidy Khlaaf, Girish Sastry, Pamela Mishkin, Brooke Chan, Scott Gray, Nick Ryder, Mikhail Pavlov, Alethea Power, Lukasz Kaiser, Mohammad Bavarian, Clemens Winter, Philippe Tillet, Felipe Petroski Such, Dave Cummings, Matthias Plappert, Fotios Chantzis, Elizabeth Barnes, Ariel Herbert-Voss, William Hebgen Guss, Alex Nichol, Alex Paino, Nikolas Tezak, Jie Tang, Igor Babuschkin, Suchir Balaji, Shantanu Jain, William Saunders, Christopher Hesse, Andrew N. Carr, Jan Leike, Josh Achiam, Vedant Misra, Evan Morikawa, Alec Radford, Matthew Knight, Miles Brundage, Mira Murati, Katie Mayer, Peter Welinder, Bob McGrew, Dario Amodei, Sam McCandlish, Ilya Sutskever, and Wojciech Zaremba.
2021.
Evaluating Large Language Models Trained on Code.

\bibitem[5]{radford2018improving}
Alec Radford, Karthik Narasimhan, Tim Salimans, Ilya Sutskever, and others.
2018.
Improving Language Understanding by Generative Pre-training.
OpenAI. Preprint, pages 1--12.

\bibitem[6]{raji2019actionable}
Inioluwa Raji and Joy Buolamwini.
 2019.
  Actionable Auditing: Investigating the Impact of Publicly Naming Biased Performance Results of Commercial AI Products.
 In \textit{Proceedings of the 2019 AAAI/ACM Conference on AI, Ethics, and Society (AIES '19)}, pages 429--435. ACM. 
 DOI: 10.1145/3306618.3314244.

\bibitem[7]{vaswani2017attention}
Ashish Vaswani, Noam Shazeer, Niki Parmar, Jakob Uszkoreit, Llion Jones, Aidan Gomez, Łukasz Kaiser, and Illia Polosukhin.
 2017.
Attention is All You Need.
In \textit{Advances in Neural Information Processing Systems (NeurIPS 2017)}. Curran Associates, Inc.

\bibitem[8]{yudkowsky2008artificial}
Eliezer Yudkowsky.
 2008.
Artificial Intelligence as a Positive and Negative Factor in Global Risk.
 In Nick Bostrom and Milan M. Cirkovic (eds.), \textit{Global Catastrophic Risks}, pages 308--345. Oxford University Press.

\bibitem[9]{zhao2017men}
Jieyu Zhao, Tianlu Wang, Mark Yatskar, Vicente Ordonez, and Kai-Wei Chang.
 2017.
Men Also Like Shopping: Reducing Gender Bias Amplification using Corpus-level Constraints.
In \textit{Proceedings of the 2017 Conference on Empirical Methods in Natural Language Processing}, pages 2979--2989. Association for Computational Linguistics.

\bibitem[10]{wang2019glue}
Alex Wang, Amanpreet Singh, Julian Michael, Felix Hill, Omer Levy, and Samuel R. Bowman.
2019.
GLUE: A Multi-Task Benchmark and Analysis Platform for Natural Language Understanding.

\bibitem[11]{wang2020superglue}
Alex Wang, Yada Pruksachatkun, Nikita Nangia, Amanpreet Singh, Julian Michael, Felix Hill, Omer Levy, and Samuel R. Bowman.
2020.
SuperGLUE: A Stickier Benchmark for General-Purpose Language Understanding Systems.

\bibitem[12]{hendrycks2021measuring}
Dan Hendrycks, Collin Burns, Steven Basart, Andy Zou, Mantas Mazeika, Dawn Song, and Jacob Steinhardt.
 2021.
  Measuring Massive Multitask Language Understanding.

\bibitem[13]{openllmleaderboard}
Open LLM Leaderboard.
 n.d.
  Open LLM Leaderboard - a Hugging Face Space by open-llm-leaderboard-old.
 Retrieved from \url{https://huggingface.co/spaces/open-llm-leaderboard-old/open_llm_leaderboard}.

\bibitem[14]{liang2023holistic}
Percy Liang, Rishi Bommasani, Tony Lee, Dimitris Tsipras, Dilara Soylu, Michihiro Yasunaga, Yian Zhang, Deepak Narayanan, Yuhuai Wu, Ananya Kumar, Benjamin Newman, Binhang Yuan, Bobby Yan, Ce Zhang, Christian Cosgrove, Christopher D. Manning, Christopher Ré, Diana Acosta-Navas, Drew A. Hudson, Eric Zelikman, Esin Durmus, Faisal Ladhak, Frieda Rong, Hongyu Ren, Huaxiu Yao, Jue Wang, Keshav Santhanam, Laurel Orr, Lucia Zheng, Mert Yuksekgonul, Mirac Suzgun, Nathan Kim, Neel Guha, Niladri Chatterji, Omar Khattab, Peter Henderson, Qian Huang, Ryan Chi, Sang Michael Xie, Shibani Santurkar, Surya Ganguli, Tatsunori Hashimoto, Thomas Icard, Tianyi Zhang, Vishrav Chaudhary, William Wang, Xuechen Li, Yifan Mai, Yuhui Zhang, and Yuta Koreeda. 2023.
Holistic Evaluation of Language Models.

\bibitem[15]{yan2024taskspecific}
Ziyou Yan.
March 2024.
Task-Specific LLM Evals that Do \& Don't Work.
eugeneyan.com. Retrieved from \url{https://eugeneyan.com/writing/evals/}.

\bibitem[16]{zellers2019hellaswag}
Rowan Zellers, Ari Holtzman, Yonatan Bisk, Ali Farhadi, and Yejin Choi.
2019.
HellaSwag: Can a Machine Really Finish Your Sentence?.

\bibitem[17]{kiela2021dynabench}
Douwe Kiela, Max Bartolo, Yixin Nie, Divyansh Kaushik, Atticus Geiger, Zhengxuan Wu, Bertie Vidgen, Grusha Prasad, Amanpreet Singh, Pratik Ringshia, Zhiyi Ma, Tristan Thrush, Sebastian Riedel, Zeerak Waseem, Pontus Stenetorp, Robin Jia, Mohit Bansal, Christopher Potts, and Adina Williams.
 2021.
Dynabench: Rethinking Benchmarking in NLP.

\bibitem[18]{dubois2024lengthcontrolled}
Yann Dubois, Balázs Galambosi, Percy Liang, and Tatsunori B. Hashimoto.
 2024.
  Length-Controlled AlpacaEval: A Simple Way to Debias Automatic Evaluators.

\bibitem[19]{papineni2002bleu}
Kishore Papineni, Salim Roukos, Todd Ward, and Wei-Jing Zhu.
 2002.
  BLEU: A Method for Automatic Evaluation of Machine Translation.
 In \textit{Proceedings of the 40th Annual Meeting on Association for Computational Linguistics}, pages 311--318. Association for Computational Linguistics.

\bibitem[20]{rajpurkar2018know}
Pranav Rajpurkar, Robin Jia, and Percy Liang.
 2018.
  Know What You Don’t Know: Unanswerable Questions for SQuAD.
 In \textit{Proceedings of the 2018 Conference on Empirical Methods in Natural Language Processing (EMNLP)}. Retrieved from \url{https://arxiv.org/abs/1806.03822}.

\bibitem[21]{alzahrani2024benchmarks}
Norah Alzahrani, Hisham Abdullah Alyahya, Yazeed Alnumay, Sultan Alrashed, Shaykhah Alsubaie, Yusef Almushaykeh, Faisal Mirza, Nouf Alotaibi, Nora Altwairesh, Areeb Alowisheq, M. Saiful Bari, and Haidar Khan.
 2024.
  When Benchmarks are Targets: Revealing the Sensitivity of Large Language Model Leaderboards.


\bibitem[22]{wallace2019universal} Eric Wallace, Shi Feng, Nikhil Kandpal, Matt Gardner, and Sameer Singh.  2019.   Universal Adversarial Triggers for Attacking and Analyzing NLP.  In Proceedings of the 2019 Conference on Empirical Methods in Natural Language Processing and the 9th International Joint Conference on Natural Language Processing (EMNLP-ĲCNLP), pages 2153–2162. Association for Computational Linguistics.


\bibitem[23]{bowman2021fixing} Samuel R. Bowman and George E. Dahl.  2021.   What Will it Take to Fix Benchmarking in Natural Language Understanding?.

\bibitem[24]{dagan2006pascal} Ido Dagan.  2006.   The PASCAL Recognising Textual Entailment Challenge.  In Machine Learning Challenges. Evaluating Predictive Uncertainty, Visual Object Classification, and Recognising Textual Entailment, pages 177–190. Springer Berlin Heidelberg.

\bibitem[25]{paperno2016lambada} Denis Paperno, Germán Kruszewski, Angeliki Lazaridou, Quan Ngoc Pham, Raffaella Bernardi, Sandro Pezzelle, Marco Baroni, Gemma Boleda, and Raquel Fernández.  2016.   The LAMBADA dataset: Word prediction requiring a broad discourse context.

\bibitem[26]{carlini2019secret} Nicholas Carlini, Chang Liu, Úlfar Erlingsson, Jernej Kos, and Dawn Song.  2019.   The Secret Sharer: Evaluating and Testing Unintended Memorization in Neural Networks.

\bibitem[27]{nie2020adversarial} Yixin Nie, Adina Williams, Emily Dinan, Mohit Bansal, Jason Weston, and Douwe Kiela.  2020.   Adversarial NLI: A New Benchmark for Natural Language Understanding.

\bibitem[28]{kaushik2020learning} Divyansh Kaushik, Eduard Hovy, and Zachary C. Lipton.  2020.   Learning the Difference that Makes a Difference with Counterfactually-Augmented Data.

\bibitem[29]{gururangan2018annotation} Suchin Gururangan, Swabha Swayamdipta, Omer Levy, Roy Schwartz, Samuel R. Bowman, and Noah A. Smith.  2018.   Annotation Artifacts in Natural Language Inference Data.

\bibitem[30]{lin2004rouge} Chin-Yew Lin.  2004.   ROUGE: A Package for Automatic Evaluation of Summaries.  In Text Summarization Branches Out, pages 74–81. Association for Computational Linguistics.

\bibitem[31]{banerjee2005meteor} Satanjeev Banerjee and Alon Lavie.  2005.   METEOR: An Automatic Metric for MT Evaluation with Improved Correlation with Human Judgments.  In Proceedings of the ACL Workshop on Intrinsic and Extrinsic Evaluation Measures for Machine Translation and/or Summarization, pages 65–72. Association for Computational Linguistics.

\bibitem[32]{dodge2021documenting} Jesse Dodge, Maarten Sap, Ana Marasović, William Agnew, Gabriel Ilharco, Dirk Groeneveld, Margaret Mitchell, and Matt Gardner.  2021.   Documenting Large Webtext Corpora: A Case Study on the Colossal Clean Crawled Corpus.

\bibitem[33]{olivier2023how} Raphael Olivier and Bhiksha Raj.  2023.   How many perturbations break this model? Evaluating robustness beyond adversarial accuracy.

\bibitem[34]{balaji2019instance} Yogesh Balaji, Tom Goldstein, and Judy Hoffman.  2019.   Instance Adaptive Adversarial Training: Improved Accuracy Tradeoffs in Neural Nets.

\bibitem[35]{xhonneux2024efficient} Sophie Xhonneux, Alessandro Sordoni, Stephan Günnemann, Gauthier Gidel, and Leo Schwinn.  2024.   Efficient Adversarial Training in LLMs with Continuous Attacks.

\bibitem[36]{alzantot2018generating} Moustafa Alzantot, Yash Sharma, Ahmed Elgohary, Bo-Jhang Ho, Mani Srivastava, and Kai-Wei Chang.  2018.   Generating Natural Language Adversarial Examples.

\bibitem[37]{ilyas2019adversarial} Andrew Ilyas, Shibani Santurkar, Dimitris Tsipras, Logan Engstrom, Brandon Tran, and Aleksander Madry.  2019.   Adversarial Examples Are Not Bugs, They Are Features.

\bibitem[38]{kale2023provenance} Aviral Kale, Truc Nguyen, Jack Harris, Chang Li, Jiayi Zhang, and Xiaoli Ma.  2023.   Provenance Documentation to Enable Explainable and Trustworthy AI: A Literature Review.  Data Intelligence, 5(1):139–162.

\bibitem[39]{oren2023proving} Yanai Oren, Nicolas Meister, Niladri Chatterji, Firoj Ladhak, and Tatsunori B. Hashimoto.  2023.   Proving Test Set Contamination in Black Box Language Models.  arXiv [Cs.CL]. Retrieved from http://arxiv.org/abs/2310.17623.

\bibitem[40]{balloccu2024leak} Silvio Balloccu, Petra Schmidtová, Matúš Lango, and Ondřej Dušek.  2024.   Leak, Cheat, Repeat: Data Contamination and Evaluation Malpractices in Closed-Source LLMs.  arXiv [Cs.CL]. Retrieved from http://arxiv.org/abs/2402.03927.

\bibitem[41]{dubois2024length} Yann Dubois, Balázs Galambosi, Percy Liang, and Tatsunori B. Hashimoto.  2024.   Length-Controlled AlpacaEval: A Simple Way to Debias Automatic Evaluators.  arXiv preprint arXiv:2404.04475.

\bibitem[42]{li2024arena} Tianle Li, Wei-Lin Chiang, Evan Frick, Lisa Dunlap, Banghua Zhu, Joseph E. Gonzalez, and Ion Stoica.  2024.   From Live Data to High-Quality Benchmarks: The Arena-Hard Pipeline.

\bibitem[43]{li2023alpacaeval} Xuechen Li, Tianyi Zhang, Yann Dubois, Rohan Taori, Ishaan Gulrajani, Carlos Guestrin, Percy Liang, and Tatsunori B. Hashimoto.  2023.   AlpacaEval: An Automatic Evaluator of Instruction-Following Models.

\bibitem[44]{white2024livebench} Colin White, Samuel Dooley, Matthew Roberts, Ameya Pal, Benjamin Feuer, Sarthak Jain, et al.  2024.   LiveBench: A Challenging, Contamination-Free LLM Benchmark.  arXiv preprint arXiv:2406.19314.

\bibitem[45]{chiang2024chatbot} Wei-Lin Chiang, Lianmin Zheng, Ying Sheng, Anastasios Nikolas Angelopoulos, Tianle Li, Dacheng Li, Hao Zhang, Banghua Zhu, Michael Jordan, Joseph E. Gonzalez, et al.  2024.   Chatbot Arena: An Open Platform for Evaluating LLMs by Human Preference.  arXiv preprint arXiv:2403.04132.

\bibitem[45]{zheng2024judging} Lianmin Zheng, Wei-Lin Chiang, Ying Sheng, Siyuan Zhuang, Zhanghao Wu, Yonghao Zhuang, Zi Lin, Zhuohan Li, Dacheng Li, Eric Xing, et al.  2024.   Judging LLM-as-a-Judge with MT-Bench and Chatbot Arena.  Advances in Neural Information Processing Systems, 36.

\bibitem[46]{clark2021all} Elizabeth Clark, Tal August, Sarah Serrano, Nate Haduong, Suchin Gururangan, and Noah A. Smith.  2021.   All That's 'Human' Is Not Gold: Evaluating Human Evaluation of Generated Text.  arXiv preprint arXiv:2107.00061.

\bibitem[47]{chiang2023can} Chung-Hsuan Chiang and Hung-Yi Lee.  2023.   Can Large Language Models Be an Alternative to Human Evaluations?  arXiv [Cs.CL]. Retrieved from http://arxiv.org/abs/2305.01937.

\bibitem[48]{touvron2023llama} Hugo Touvron, et al.  2023.   LLaMA: Open and Efficient Foundation Language Models.  arXiv preprint arXiv:2302.13971.

\bibitem[49]{openai2024gpt4} OpenAI, Josh Achiam, Steven Adler, Sandhini Agarwal, Lama Ahmad, Ilge Akkaya, Florencia Leoni Aleman, Diogo Almeida, Janko Altenschmidt, Sam Altman, Shyamal Anadkat, et al.  2024.   GPT-4 Technical Report.

\bibitem[50]{bai2023qwen} Inze Bai, Shuai Bai, Yunfei Chu, Zeyu Cui, Kai Dang, Xiaodong Deng, Yang Fan, Wenbin Ge, Yu Han, Fei Huang, Binyuan Hui, Luo Ji, Mei Li, Junyang Lin, Runji Lin, Dayiheng Liu, Gao Liu, Chengqiang Lu, Keming Lu, Jianxin Ma, Rui Men, Xingzhang Ren, Xuancheng Ren, Chuanqi Tan, Sinan Tan, Jianhong Tu, Peng Wang, Shĳie Wang, Wei Wang, Shengguang Wu, Benfeng Xu, Jin Xu, An Yang, Hao Yang, Jian Yang, Shusheng Yang, Yang Yao, Bowen Yu, Hongyi Yuan, Zheng Yuan, Jianwei Zhang, Xingxuan Zhang, Yichang Zhang, Zhenru Zhang, Chang Zhou, Jingren Zhou, Xiaohuan Zhou, and Tianhang Zhu.  2023.   Qwen Technical Report.

\bibitem[51]{strathern1997audit}
Marilyn Strathern.
1997.
‘Improving ratings’: audit in the British University system.
\textit{European Review}, 5(3), 305–321.
doi:10.1002/(SICI)1234-981X(199707)5:3<305::AID-EURO184>3.0.CO;2-4.

\bibitem[52]{sokolova2006beyond}
Marina Sokolova, Nathalie Japkowicz, and Stan Szpakowicz.
2006.
Beyond Accuracy, F-Score and ROC: A Family of Discriminant Measures for Performance Evaluation.
In \textit{AI 2006: Advances in Artificial Intelligence, Lecture Notes in Computer Science}, Vol. 4304, pages 1015-1021. doi:10.1007/11941439\_114.

\bibitem[53]{lin2022truthfulqa}
Stephanie Lin, Jacob Hilton, and Owain Evans.
2022.
TruthfulQA: Measuring How Models Mimic Human Falsehoods.
Available at \url{https://arxiv.org/abs/2109.07958}.

\bibitem[54]{srivastava2023beyond}
Aarohi Srivastava, Abhinav Rastogi, Abhishek Rao, Abu Awal Md Shoeb, Abubakar Abid, Adam Fisch, Adam R. Brown, Adam Santoro, Aditya Gupta, Adrià Garriga-Alonso, et al.
2023.
Beyond the Imitation Game: Quantifying and Extrapolating the Capabilities of Language Models.
Available at \url{https://arxiv.org/abs/2206.04615}.

\bibitem[55]{zhu2020freelb}
Chen Zhu, Yu Cheng, Zhe Gan, Siqi Sun, Tom Goldstein, and Jingjing Liu.
2020.
FreeLB: Enhanced Adversarial Training for Natural Language Understanding.
Available at \url{https://arxiv.org/abs/1909.11764}.

\bibitem[56]{williams2018broadcoverage}
Adina Williams, Nikita Nangia, and Samuel R. Bowman.
2018.
A Broad-Coverage Challenge Corpus for Sentence Understanding through Inference.
Available at \url{https://arxiv.org/abs/1704.05426}.

\bibitem[57]{zhuge2024agentasajudgeevaluateagentsagents}
Mingchen Zhuge, Changsheng Zhao, Dylan Ashley, Wenyi Wang, Dmitrii Khizbullin, Yunyang Xiong, Zechun Liu, Ernie Chang, Raghuraman Krishnamoorthi, Yuandong Tian, Yangyang Shi, Vikas Chandra, Jürgen Schmidhuber.
2024.
Agent-as-a-Judge: Evaluate Agents with Agents.
Available at \url{https://arxiv.org/abs/2410.10934}.



\end{thebibliography}

\end{document}